\documentclass[10pt,twocolumn,letterpaper]{article}

\usepackage{cvpr}
\usepackage{times}
\usepackage{epsfig}
\usepackage{graphicx}
\usepackage{amsmath}
\usepackage{amssymb}

\usepackage[pagebackref=true,breaklinks=true,letterpaper=true,colorlinks,bookmarks=false]{hyperref}

 \cvprfinalcopy 

\def\cvprPaperID{1638} 

\ifcvprfinal\fi
\begin{document}

\title{DeMeshNet: Blind Face Inpainting for Deep MeshFace Verification}

\author{Shu Zhang\quad Ran He\quad Tieniu Tan\\
National Laboratory of Pattern Recognition, CASIA\\
Center for Research on Intelligent Perception and Computing, CASIA\\
{\tt\small \{shu.zhang,rhe,tnt\}@nlpr.ia.ac.cn}
}

\maketitle

\begin{abstract}
MeshFace photos have been widely used in many Chinese business organizations to protect ID face photos from being misused. The occlusions incurred by random meshes severely degenerate the performance of face verification systems, which raises the MeshFace verification problem between MeshFace and daily photos. Previous methods cast this problem as a typical low-level vision problem, \ie blind inpainting. They recover perceptually pleasing clear ID photos from MeshFaces by enforcing pixel level similarity between the recovered ID images and the ground-truth clear ID images and then perform face verification on them.

Essentially, face verification is conducted on a compact feature space rather than the image pixel space. Therefore, this paper argues that pixel level similarity and feature level similarity jointly offer the key to improve the verification performance. Based on this insight, we offer a novel feature oriented blind face inpainting framework. Specifically, we implement this by establishing a novel DeMeshNet, which consists of three parts. The first part addresses blind inpainting of the MeshFaces by implicitly exploiting extra supervision from the occlusion position to enforce pixel level similarity. The second part explicitly enforces a feature level similarity in the compact feature space, which can explore informative supervision from the feature space to produce better inpainting results for verification. The last part copes with face alignment within the net via a customized spatial transformer module when extracting deep facial features. All the three parts are implemented within an end-to-end network that facilitates efficient optimization. Extensive experiments on two MeshFace datasets demonstrate the effectiveness of the proposed DeMeshNet as well as the insight of this paper.
\end{abstract}


\section{Introduction}\label{sec:intro}





Benefitting from recent advancements in deep representation learning, there have been remarkable improvements in deep face recognition (verification in particular)~\cite{schroff2015facenet,sun2014deep,taigman2014deepface}. In real life applications, face verification between ID photos and daily photos (FVBID)~\cite{zhou2015naive} is gaining traction because it uses a face image from an ID photo as gallery and thus does not require the probe to be registred in advance.



\begin{figure}
  \centering
    \includegraphics[scale=0.23]{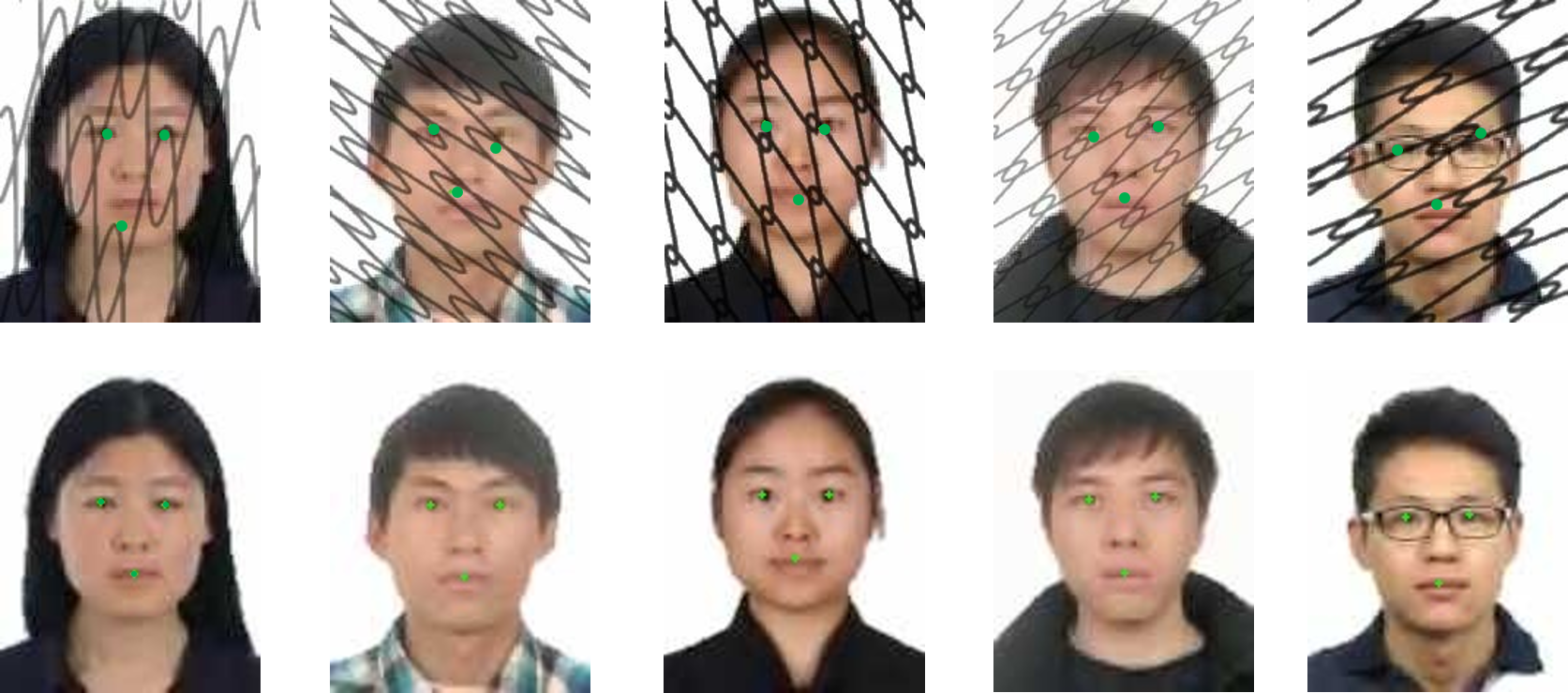}
    \caption{MeshFaces (first row) refer to the ID photos corrupted by randomly generated mesh-like lines or watermarks. The corruptions significantly degenerate the performance of facial landmark detection and facial feature extraction, thus leading to poor verification accuracy.}
    \label{fig:spn} 
\end{figure}


When FVBID is applied to real-world scenarios, such as automated custom control and VIP recognition in commercial banks, the ID photo in an identity card may potentially be misused or illegally distributed. Therefore, ID photos are often deliberately corrupted by mesh-like lines or watermarks for privacy protection when used by some business organizations, \eg banks and hotels. For convenience, we denote this type of corrupted ID photo as MeshFace. As shown in Fig.~\ref{fig:spn}, MeshFaces incur catastrophic influence to face recognition systems~\cite{Burgos-Artizzu_2013_ICCV,he}. Directly verifying MeshFaces against daily photos leads to very poor accuracy~\cite{7550058}. Therefore, these corruptions raise a novel and challenging problem called MeshFace verification which deals with face verification between MeshFaces and daily photos.

Some efforts have been made to address this challenging problem. Zhang et al.~\cite{7550058} propose a multi-task residual learning CNN for this problem. They propose to learn a non-linear transformation with SRCNN~\cite{dong2014learning} based architecture to recover clear ID photos from MeshFaces. Then, the recovered clear ID photos are used for face verification. They treat the recovery of clear ID photos from MeshFaces as \textit{blind face inpainting} because the position of corruptions is unknown during testing phase. In a related vein of research, many contemporary works~\cite{mao2016image,Pathak_2016_CVPR,ren2015shepard,xie2012image} have shown that CNN is very effective in solving hole filling (non-blind inpainting) problems~\cite{criminisi2004region} because this data-driven learning method can exploit the structure of natural images to predict occluded parts.
\begin{figure}
  \centering
    \includegraphics[scale=0.23]{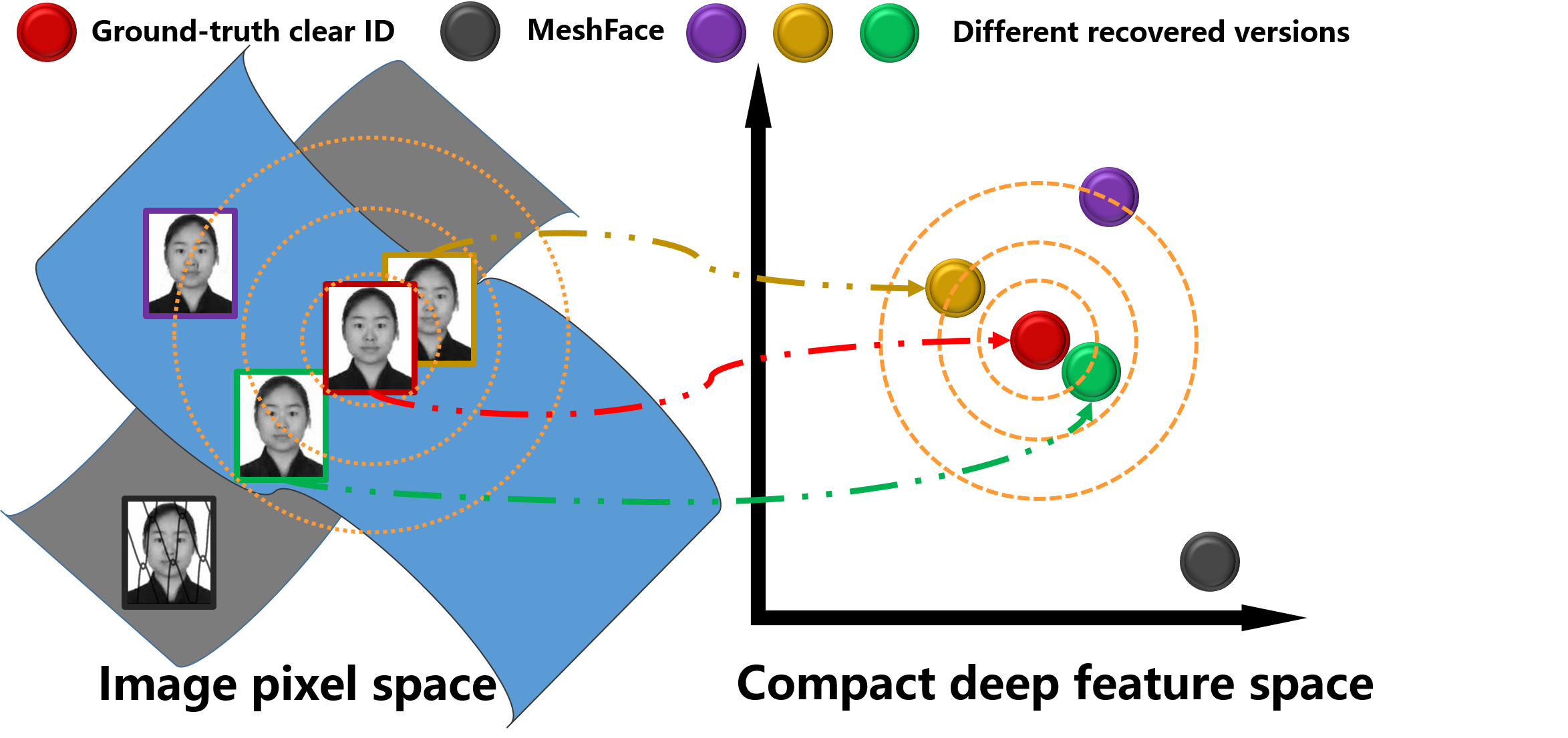}
    \caption{
Recovered ID with higher PSNR may be further away from the ground-truth clear ID in the compact deep feature space. For instance, the green and yellow sample. Best viewed in color.}
    \label{fig:insight} 
\end{figure}

Improved verification performance is observed after blind face inpainting in~\cite{7550058} because using an occlusion free image will greatly improve the accuracy of face detection and alignment. However, in their work, the performance gap between using their inpainted ID and the clear ID is still very large. On one hand, this is because SRCNN is less powerful in modeling corruption distributions and recovering the exact image content. As a result, the difference in image content between the recovered ID photos and the ground-truth clear ID photos are too large (as illustrated by the red and the purple sample in Fig.\ref{fig:insight}).

\begin{figure*}
  \centering
    \includegraphics[scale=0.43]{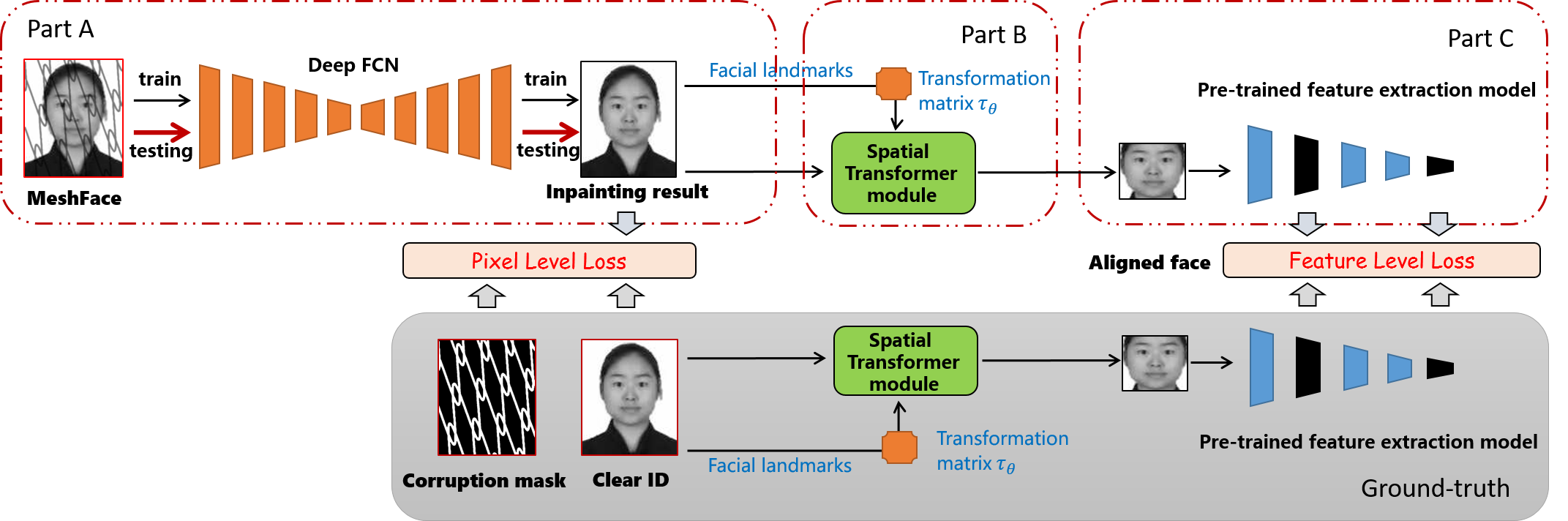}
    \caption{Conceptual diagram of DeMeshNet. Black solid lines stand for the training phase and red solid lines represent the testing phase. Note that the corruption mask and clear ID are only used in the training phase to provide ground-truth for the pixel and feature level loss. Feature extraction model is pre-trained and has fixed parameters during training. In the testing phase, DeMeshNet takes a MeshFace as input and produce an inpainted photo with the learned deep FCN.}
    \label{fig:overview} 
\end{figure*}




On the other hand, existing works often assume that using recovered ID photos with higher PSNR are more likely to achieve better verification performance~\cite{7550058}. But in fact, face similarity is compared in a compact feature space rather the image pixel space. Moreover, it has been shown that CNN can be potentially `fooled' by adding even a tiny amount of noise to the original input~\cite{goodfellow2014explaining,nguyen2015deep,szegedy2013intriguing}. That is, when facial features are extracted by a CNN, two perceptually indistinguishable face images (\eg the ground-truth clear ID and its recovered version) may still have very large feature level differences (as shown by the green and yellow sample in Fig.\ref{fig:insight}). It is a common belief that a large intra-class feature distance will generally deteriorate face verification performance. Therefore, this suggests that treating blind face inpainting as a typical low-level vision problem by only enforcing the pixel level similarity can hardly guarantee an improvement in verification performance.


To address the aforementioned problems, we present DeMeshNet to take verification performance into account when dealing with the blind face inpainting problem. DeMeshNet is trained on a large scale dataset of  MeshFace/clear ID photo pairs to learn a non-linear transformation to recover clear ID photos from MeshFaces. Note that DeMeshNet aims to improve the MeshFace verification performance rather than simply to recover perceptually pleasing clear ID photos. Therefore, we refer to DeMeshNet as a feature oriented blind face inpainting framework. We briefly introduce each part of the DeMeshNet and our contributions in the following paragraphs.



In the first part, we enforce the \textit{pixel level similarity} to explore the structure of face images so as to recover uncorrupted ID photos from MeshFaces. Specifically, we propose to adopt a fully convolutional network (FCN)~\cite{long2015fully} with a weighted Euclidean loss to minimize the pixel differences of the ground-truth clear ID/recovered ID photo pairs. Extra supervision from corruption positions is further exploited to accurately model the corruption distribution.


In the second part, we enforce a \textit{feature level similarity} between the ground-truth clear ID photo and the recovered ID photo pairs in a compact deep feature space. This feature space is spanned by a pre-trained CNN, and distance in it directly corresponds to a measure of face similarity. This part will force the inpainted image to have a smaller distance to the ground-truth clear ID in the deep feature space, which will in turn facilitate accurate verification. Moreover, we propose to measure the feature level similarity with the reverse Huber loss function so that the feature level similarity can be more efficiently optimized when the differences are very small.

In the third part, we employ a \textit{customized spatial transformer module}~\cite{jaderberg2015spatial} to align and crop the face region for accurate feature extraction within the network. It is essential to take alignment into account because the MeshFace which is the input of DeMeshNet and the aligned face which is the input of the feature extraction sub-net are different in sizes, scales and orientations (as shown in Fig.~\ref{fig:overview}).

All the three parts are implemented within an end-to-end network that facilitates efficient optimization with gradient back propagation. Extensive experimental results on two MeshFace datasets demonstrate that DeMeshNet achieves the best verification accuracy and outperforms previous work by a large margin.
Furthermore, we thoroughly evaluate different configurations of DeMeshNet to gain insight into the factors for such significant improvements.

\section{Approach}

\subsection{Overview}
In this section, we present an overview of the proposed DeMeshNet. We cast the proposed feature oriented blind face inpainting problem as a dense regression problem, which aims to regress a perceptually pleasing and verification favorable clear ID photo from a MeshFace $X$. For convenience, we refer to the ground-truth clear ID photo as the \textit{target}, termed as $Y$  and the recovered ID photo from our blind face inpainting model $\psi$ as the \textit{prediction}, termed as $\psi ({X})$. The prediction will be used for face verification against clear daily photos.

We model the highly non-linear function for dense regression as a FCN as illustrated in part A of Fig.~\ref{fig:overview}. Part B shows the customized spatial transformer module. The following part is a pre-trained CNN which is utilized to compute the feature representation of the aligned face region. It should be noted that parameters in both the spatial transformer module and the pre-trained CNN are fixed during training. The learnable parameters in the FCN are optimized through minimizing a unified loss function that jointly models the pixel and feature level similarities between the prediction and the target pairs. No identity information is needed to train such a blind inpainting network.

The pixel level loss helps to obtain perceptually pleasing inpainting results and serves as a means to capture the distribution difference between actual face texture and mesh-like corruptions. And the feature level loss explores supervision in a compact feature space to provide regularizations to the network training. Thus, the network's prediction will not only have similar appearance but also have similar feature representation to that of the target. Specifically, we develop a weighted Euclidean loss to model the pixel level similarity and employ a reverse Huber function~\cite{laina2016deeper} to characterize the feature level loss on the spatial transformed face region. Combining the pixel level loss and the spatial transformed feature level loss, we define the unified loss function for DeMeshNet as follows:
\begin{equation}
\begin{split}
&{L_i} = l_{pixel} + {l_{feature}} = \\
&||\psi ({X_i} ) - {Y_i}||_F^2 + \lambda_1 ||{M_i} \odot (\psi ({X_i} ) - {Y_i})||_F^2+\\
 &\lambda_2\sum\limits_{j=1}^2 {RH({\phi _j}(\psi ({ST(X_i)})) - {\phi _j}({ST(Y_i)})} {\kern 1pt} {\kern 1pt} )
\end{split}\label{eq:first}
\end{equation}
where $ST$ denotes the spatial transformation implementation that samples an aligned $128\times 128$ face region from the original input solely based on the positions of facial
s, $RH$ is the reverse Huber function, $\lambda_1$ and $\lambda_2$ are the balance parameters which are empirically set to $1$ throughout the paper. We postpone explanations of other symbols to later sections when we meet them.

For simplicity, we omit the regularization term on the parameters of FCN (weight decay) in Equation~\ref{eq:first}, which is used to reduce overfitting when optimizing our network. The objective function can be efficiently optimized by gradient back propagation in an end-to-end manner. We will elaborate each of the three parts in the following subsections.

\subsection{Pixel Level Regression Network}

\subsubsection{Pixel Level Loss}Blind face inpainting is naturally characterized as a pixel-wise regression problem. In the first part of DeMeshNet, it learns a highly non-linear transformation by optimizing a well-designed pixel level loss. For blind face inpainting, although positions of the corruption are not provided during the testing phase, we can still make use of this information when training DeMeshNet. Specifically, we propose to implicitly exploit this extra supervision by introducing a weighted Euclidean loss function as below:
\begin{equation}
{l_{pixel}} =||\psi ({X_i} ) - {Y_i}||_F^2 + \lambda ||{M_i} \odot (\psi ({X_i} ) - {Y_i})||_F^2
\end{equation}

where $X_i$, $Y_i$ and $\psi ({X_i})$ are a corrupted input, target and prediction, respectively. $M_i$ is the binary mask with a value of $1$ indicating the pixel is corrupted and a value of $0$ otherwise. $\odot$ is the element-wise product operation. Therefore, the second term in the loss function only measures the Euclidean loss on corrupted areas, emphasizing losses on those areas with a weighting parameter $\lambda$. This loss function helps the network to learn the distributions of corrupted pixels better by exploiting extra supervision from the corruption positions. Experimental results demonstrate that it also helps to generate predictions with higher PSNR.

\subsubsection{Network Architecture} Since FCN has achieved outstanding performance in dense prediction tasks like depth prediction~\cite{eigen2014depth} and semantic segmentation~\cite{long2015fully}, it motivates us to use FCN as the non-linear function to improve the blind face inpainting performance.

The main difference between FCN and the architecture in~\cite{7550058} is the introduction of down-sampling and up-sampling layers in FCN. This simple adjustment has enabled FCN to admit much deeper layers and to expand the receptive fields with the same amount of computational cost. The expanded receptive fields are critical to blind inpainting as it can enclose more contextual information for identifying the corrupted areas.

In this work, we use the network architecture of SegNet as proposed in~\cite{badrinarayanan2015segnet}.
 It is feasible to adopt other architectures such as ResidualNet~\cite{He_2016_CVPR} and Deconvolution Network~\cite{Noh_2015_ICCV}, but this is beyond the scope of this paper. The input and output to the network are MeshFaces and clear ID photos respectively. Gray scale images of size $220 \times 178$ are used for input and output throughout the paper.

\subsection{Feature Level Regression Network}
\subsubsection{Feature Level Loss} As aforementioned in Section~\ref{sec:intro}, images with very small Euclidean distance may have large feature distance. This problem is raised in~\cite{szegedy2013intriguing} and has been shown to severely deteriorate classification performance because of the enlarged intra-class feature distance. In fact, the enlarged intra-class feature distance can also influence the verification task at hand. To improve the verification performance, it won't be enough to only enforce pixel level similarity. 
Therefore, in the second part, we explicitly enforce the target and the prediction to have a small distance in the compact feature space computed by the pre-trained CNN $\phi$.

Let $\phi _j(\psi(X_i))$ be the activations of the $j$th layer of the pre-trained face model $\phi$. We intend to improve the verification performance by minimizing the residual $r = {\phi (\psi ({x_i})) - \phi ({y_i})}$ at each position of an image. To efficiently back-propagate the errors when the residual is very small, we employ the reverse Huber loss function~\cite{laina2016deeper} to measure the feature level difference, its formulation is:
\begin{equation}
RH(r) = \left\{ \begin{array}{l}
\left| r \right| \ \ \ \ \ \ \ \ \  \left| r \right| > c\\
\frac{{r^2 + c^2}}{{2c}}  \ \ \ \ \left| r \right| \le c
\end{array} \right.
\end{equation}

The reverse Huber loss is equivalent to L1 norm when the residual $r$ is in the interval of $[-c,c]$ and equals to a transformed L2 norm otherwise. This loss experimentally works better than L2 norm.
 Note that when $c$ is smaller than $1$, the derivative of the L1 norm is greater than that of the L2 norm, which will speed up error back-propagation when the residual is very tiny. Like in~\cite{laina2016deeper}, we use a dynamic threshold for $c$. That is, in each batch-minimization step, $c$ is set to be at $20\%$ of maximum residual in that batch.

Since our objective is to improve verification performance, we impose the reverse Huber loss on the output of $eltwise\_6$ in the pre-trained model $\phi$, which is a $256$-dim feature vector used for similarity comparison. Drawing inspirations from~\cite{johnson2016perceptual}, where feature loss is used for super-resolution, we also impose the reverse Huber loss on the early layers ($conv2$ in our case) of the pre-trained face model $\phi$ to include deeper supervision~\cite{lee2014deeply}. Therefore, our final loss function for feature level regression is:
\begin{equation}
{l_{feature}} = \sum\limits_{j=1}^2 {RH({\phi _j}(\psi ({X_i})) - {\phi _j}({Y_i})} {\kern 1pt} {\kern 1pt} )
\end{equation}

Feature loss has recently been considered in the literature of super-resolution and sketch inversion for better visual performance~\cite{guccluturk2016convolutional,johnson2016perceptual,LedigTHCATTWS16}. But it should be noted that in this paper, the feature level loss is proposed from a totally different perspective. Instead of pursing a perceptually pleasing image transformation results, we want to deal with the large intra-class feature distance problem and improve verification performance.

\begin{figure}
  \centering
    \includegraphics[scale=0.23]{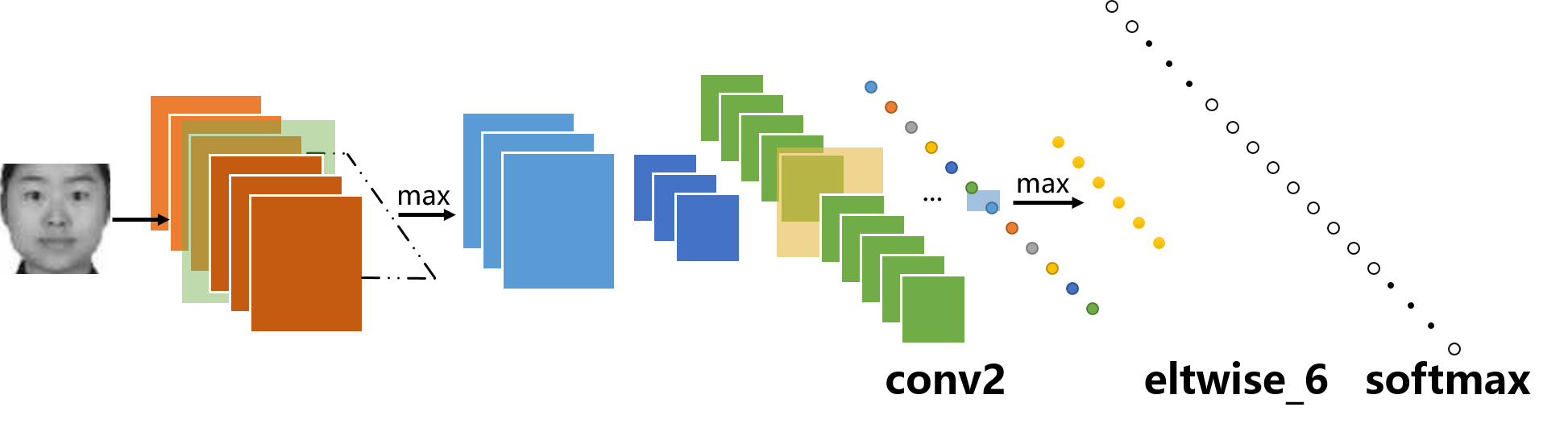}
    \caption{Schematic architecture for face feature extraction. It uses Max-Feature-Map nonlinearities instead of ReLU.}
    \label{fig:face} 
\end{figure}


\subsubsection{Pre-trained CNN for Face Feature Extraction} Both training DeMeshNet and evaluate the verification performance need a pre-trained CNN to compute the facial features for face images. We use the architecture proposed in~\cite{wu2015lightened} for facial feature extraction because it is computationally efficient in both time and space. Fig.~\ref{fig:face} briefly illustrates its architecture, which uses Max-Feature-Map nonlinearities instead of ReLU and thus can generate dense features at its output. The network takes aligned gray-scale face images of size $128 \times 128$ as input and returns a $256$-dim feature (output of $eltwise\_6$). Alignment is conducted by transforming two facial landmarks (i.e., centers of two eyes) to $(32,32)$ and $(96,32)$ with a similarity transformation.

We train a base model on the \textit{purified} MS-Celeb-1M dataset~\cite{guo2016ms} (the original data is very noisy, we purify them before use). One single base model achieves a verification accuracy of $98.80\%$ on the LFW~\cite{LFWTech} benchmark, which is very competitive among already published works~\cite{ding2015robust,wang2015face}. We further finetune the base model with triple-let loss on a large-scale ID-daily photo dataset collected from the web and use the finetuned model $\phi$ to extract compact facial features in DeMeshNet. Note that this model's parameters should stay fixed during the training of DeMeshNet because we only use it for feature computation and no identity related supervision is adopted to finetune $\phi$ while learning the blind inpainting FCN.

\subsection{Spatial Transformer For Face Alignment} Face alignment is essential for feature extraction. It helps to improve verification performance by providing a normalized input. The pre-trained model $\phi$ admits $128\times 128$ aligned face region to compute the $256$-dim feature. But DeMeshNet takes $220\times 178$ un-aligned MeshFace as input and outputs a prediction with the same size. Therefore, in order to compute the $256$-dim feature in the fully connected layer $eltwise\_6$, we must implement face alignment within the network.

\begin{figure}
  \centering
    \includegraphics[scale=0.43]{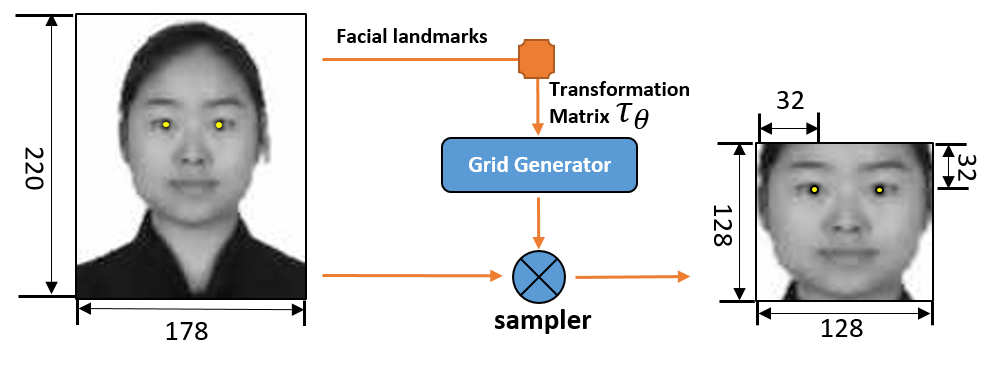}
    \caption{The customized spatial transformer module used in our model. It uses the locations of facial landmarks to calculate a transformation matrix for face alignment.}
    \label{fig:stn} 
\end{figure}

Moreover, unlike the image classification models that are trained with multi-scale natural images from ImageNet~\cite{russakovsky2015imagenet}, the pre-trained facial feature extraction model $\phi$ only takes single-scale, well-aligned face regions for training. This means that even we only compute the features from conv layers like in~\cite{guccluturk2016convolutional,johnson2016perceptual,LedigTHCATTWS16}, we will still need to align the MeshFaces first to acquire an accurate feature representation.



We incorporate a customized spatial transformer module~\cite{jaderberg2015spatial} between the pixel level regression sub-net and the feature level regression sub-net to sample an aligned $128 \times 128$ face region from the $220 \times 178$ prediction according to the facial landmarks. This procedure is illustrated in Fig.~\ref{fig:overview} and detailed in Fig.~\ref{fig:stn}.

The spatial transformer module comprises of a localization network, a grid generator and a sampler. Since the similarity transformation $\tau_\theta$ for face alignment is uniquely determined by the coordinates of two eye centers, we do not need to learn it through the localization network as in~\cite{jaderberg2015spatial}. $\tau_\theta$ is parameterized by $\theta = [a,b,1;-b,a,1]$ and can be determined with the following equation:
\begin{equation}
\left( {\begin{array}{*{20}{c}}
x_l&x_r\\
y_l&y_r\\
1&1
\end{array}} \right) = \left[ {\begin{array}{*{20}{c}}
a&b&1\\
{ - b}&a&1
\end{array}} \right]\left( {\begin{array}{*{20}{c}}
{ - 0.5}&{0.5}\\
{ - 0.5}&{ - 0.5}\\
1&1
\end{array}} \right)
\end{equation}
where $(x_l,y_l)$, $(x_r,y_l)$ and $(-0.5,-0.5)$, $(0.5,-0.5)$ are the \textit{normalized coordinates} (normalized to $[-1,1]$) of two eye centers in the original image and the aligned face image respectively.


In the forward pass, a sampling grid is firstly determined with the given $\tau_\theta$. A sampling grid is a set of points with continuous coordinates. Sampling an input image according to this sampling grid will generate a transformed output.
 By defining the output pixels to lie on a regular grid $G = {G_i}$ of pixels $G_i = (x_i^{t},y_i^{t})$, the sampling grid $\tau_\theta(G)$ is given by the point-wise transformation:
\begin{equation}
\left( \begin{array}{l}
x_i^{s}\\
y_i^{s}
\end{array} \right) = {\tau _\theta }(G) = \left[ {\begin{array}{*{20}{c}}
a&b&1\\
{ - b}&a&1
\end{array}} \right]\left( \begin{array}{l}
x_i^{t}\\
y_i^{t}
\end{array} \right)
\end{equation}
where $(x_i^{s}, y_i^{s})$ are the source coordinates in the input feature map, and $(x_i^{t}, y_i^{t})$ are the target coordinates of the regular grid. Since the coordinates in the sampling grid are continuous numbers, a bilinear kernel is applied to those positions to produce the corresponding pixel values in the output:
\begin{equation}
Q = max(0,1 - \left| {x_i^{s} - m} \right|)max(0,1 - \left| {y_i^{s} - n} \right|)
\end{equation}
\begin{equation}
P_{(x_i^{t},y_i^{t})}= \sum\limits_n^H {\sum\limits_m^W {P_{(x_n^{s},y_m^{s})}Q} }
\end{equation}
where $H$ and $W$ are the height and width of the input image respectively and $P_{(x_i^{t},y_i^{t})}$ represents the pixel value of $(x_i^{t},y_i^{t})$. To allow the feature loss defined in the last subsection to be back-propagated from the output of the spatial transformer module to the input image, we give the gradients with respect to the input image as follows:
\begin{equation}
\frac{{\partial (P(x_i^t,y_i^t))}}{{\partial (P(x_m^s,y_n^s))}} =\sum\limits_n^H {\sum\limits_m^W {Q} }\label{eq:bp}
\end{equation}

The gradients of the feature loss defined earlier can be easily flowed back to the input image using chain rule with Equation~\ref{eq:bp}. Note that the gradients with respect to the sampling grid coordinates $(x_i^{s}, y_i^{s})$ are not derived, because the transformation parameters are not learned in the customized spatial transformer module.

\begin{figure}
  \centering
    \includegraphics[scale=0.23]{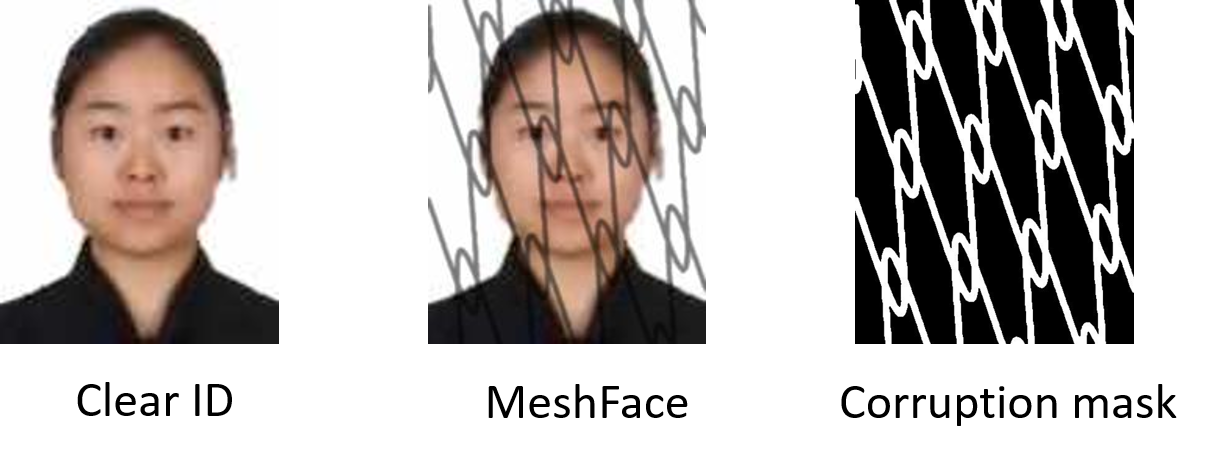}
    \caption{An image triplet sample from the training dataset~\cite{7550058}.}
    \label{fig:dataset} 
\end{figure}


\section{Experiments}

In this section, we experimentally evaluate the proposed framework. We begin by introducing the datasets for training and testing. Then we specify the baseline methods and implementation details. At last, we present detailed algorithmic evaluation, as well as comparison with other methods.

\subsection{Datasets}

\begin{figure}
  \centering
    \includegraphics[scale=0.23]{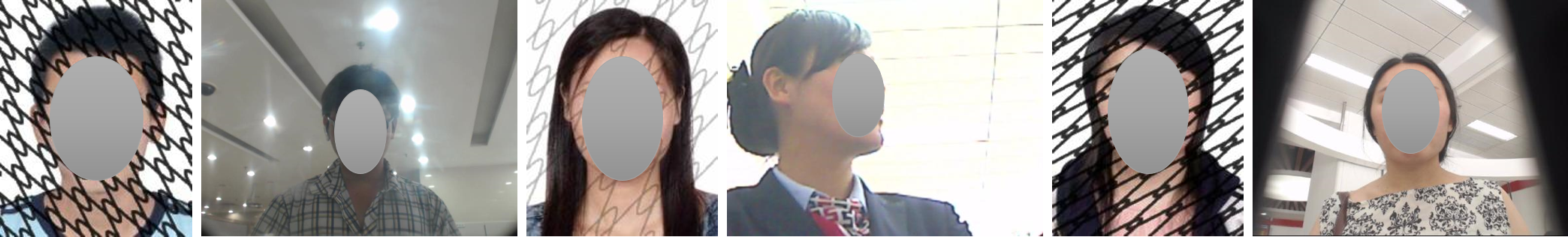}
    \caption{Sample image pairs from SV1000. Note that this dataset is very hard as the variations in illumination, pose and hair style are very significant.}
    \label{fig:sv} 
\end{figure}

All the compared models are trained on the dataset as in~\cite{7550058} that contains over $500,000$ data triplets of $11,648$ individuals. Each data triplet consists of a MeshFace, its clear version and a corruption mask (as illustrated in Fig.~\ref{fig:dataset}). Facial landmarks (two eye centers) are detected using Intraface~\cite{XiongD13} to aid the spatial transformer module. Data triplets of $400$ individuals are sampled for validation and testing ($200$ each) and all the other individuals are used for training.

\begin{figure}
  \centering
    \includegraphics[scale=0.23]{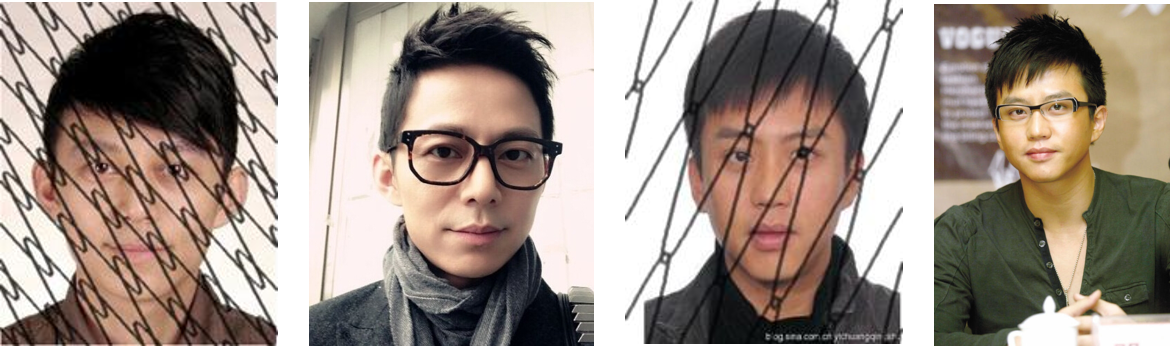}
    \caption{Sample image pairs from SYN500. Images of some Chinese celebrities are collected from the Internet.}
    \label{fig:syn} 
\end{figure}

Besides the SYN500 used in~\cite{7550058}, we collect another dataset with 1000 MeshFace/daily photo pairs from 1000 individuals named SV1000 to evaluate the MeshFace verification performance. Daily photos in SV1000 are captured under surveillance cameras. As shown in Fig.~\ref{fig:sv}, this dataset not only contains more individuals but also presents more variations in the daily photo, which makes it more challenging than SYN500 (shown in Fig.~\ref{fig:syn}).

We develop a protocol for evaluation of verification performance on these two datasets. Specifically, face comparison is conducted between all the possible recovered clear ID/daily photo pairs in the compact feature space (spanned by model $\phi$) with cosine distance. For a dataset with $N$ data pairs, $N^2$ comparisons are conducted in total. To exclude influences from metric learning methods, no supervised learning methods, \eg joint bayesian~\cite{chen2012bayesian}, are employed on the extracted features.

\subsection{Baselines and Implementation Details}

Although many algorithms have been proposed for non-blind inpainting, few have been developed to address the blind inpainting problem, even less for the blind face inpainting problem addressed in this paper. We implement the multi-task CNN (MtNet)~\cite{7550058} as a baseline. This method employs architecture that resembles the SRCNN~\cite{dong2014learning} and use multi-task learning to make use of the information of corruption position in the training phase.

To give a detailed evaluation of each part of the proposed DeMeshNet, we also compare it with various configurations. The compared configurations include FCN with Euclidean pixel level loss (FCNE), FCN with weighted pixel level loss (FCNW) and feature loss FCN without spatial transformer module (FCNF). All three configurations use FCN as the backbone for the blind face inpainting task. Both FCNE and FCNW only adopts the pixel level loss during training, but FCNW introduces an implicit supervision from the corruption mask with a weighted loss in addition to the Euclidean loss. FCNF takes both weighted pixel loss and whole-image feature level loss into consideration. But the feature level differences are only computed at the output of $conv2$, using whole-image as input to the pre-trained feature extraction network $\phi$. Like in~\cite{johnson2016perceptual}, we don't implement face alignment within the network.

All the compared models are trained on the training set with photo pairs of size $220 \times 178$, gray scale images are used in all the experiments. For all the compared network structure, training is carried out using Adam~\cite{adam} with a batch size of $30$. The learning rate is set to $10^{-4}$ initially, and decreased by a factor of $10$ each $40k$ iterations. The training process takes approximately $160k$ iterations to converge. For the proposed approach, feature level loss is computed at layer $conv2$ and $eltwise\_6$ of the pre-trained face model $\phi$. For FCNF, only $conv2$ is used for computing the feature loss as the computation of $fc$ layer $eltwise\_6$ requires the input to be of size $128\times 128$. All the MeshFace verification experiments use the pre-trained face model $\phi$ for facial feature extraction. All the experiments are conducted with the Caffe framework~\cite{jia2014caffe} on a single GTX Titan X GPU.


\begin{table*} 
\centering  
\caption{Verification performance on SYN500/SV1000 and inpainting results on the testing set. RMSE illustrates the feature distance between the recovered ID photos and the ground-truth clear ID photos, while PSNR indicates the pixel distance. Note that smaller RMSE consistently indicates better verification performance, but higher PSNR doesn't guarantee that.}  
 \begin{tabular}{|c|c|c|c|c|c|}  

     \hline
       Method&TPR@FPR=1\% &  TPR@FPR=0.1\% & TPR@FPR=0.01\%&PSNR&RMSE  \\
       \hline
            \hline
       MtNet &83.60\% / 78.80\% & 62.80\% / 57.50\%&36.80\% / 35.40\% &29.89& 55.47    \\
       Clear & \bf{98.80\%} / 88.10\%& \bf{89.40\%} / 74.30\%&\bf{67.40\% / 53.60\%}&-&0\\
       Corrupted & 47.40\% / 43.20\% & 33.80\% / 28.50\%&18.40\% / 18.20\%&20.69&112.63\\
       \hline
       FCNE & 95.20\% / 83.20\% & 79.80\% / 63.90\%&53.80\% / 43.00\% & 35.11&49.52\\
       FCNW & 95.40\% / 85.70\% & 78.40\% / 64.90\%  & 52.80\% / 44.40\%&\bf{35.31}&48.22\\
        FCNF & 95.20\% / 85.80\% & 82.80\% / 66.40\% &54.40\% / 46.30\% &  25.26& 38.19  \\
        DeMeshNet\_E & 96.60\% / 86.30\% & 83.90\% / 70.00\% &54.80\% / 46.50\% &  29.28& 35.77  \\
        \hline
        DeMeshNet & \bf{97.40\%} / 86.70\% & \bf{84.80\%} / 70.70\%&\bf{55.20}\% / 47.00\%&29.16&\bf{34.57} \\
       \hline

   \end{tabular}\label{tab:psnr}
\end{table*}

\subsection{Evaluation of Verification Results}

\begin{figure}
  \centering
    \includegraphics[scale=0.3]{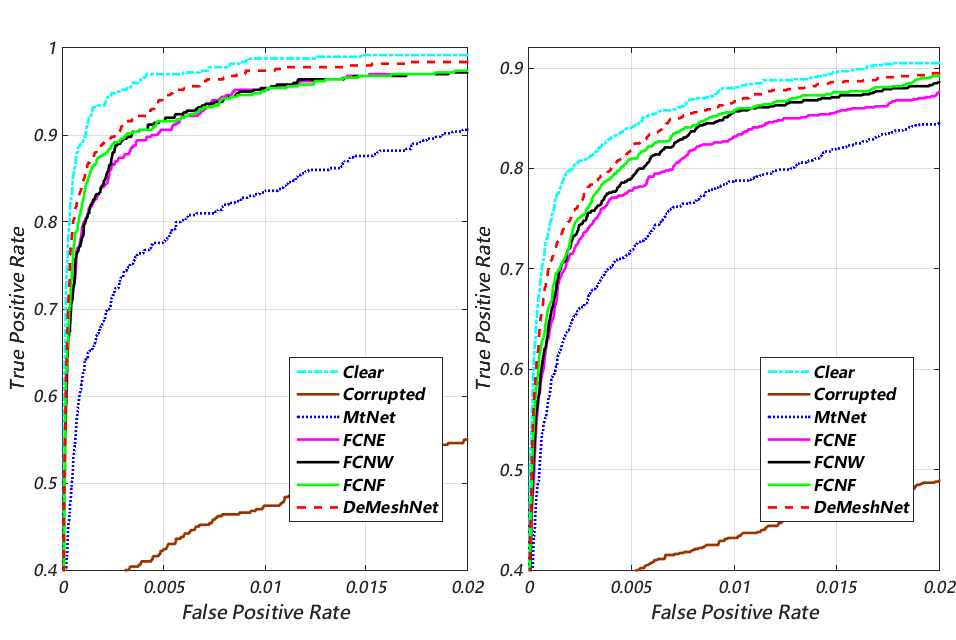}
    \caption{ROC curves for SYN500 and SV1000.}
    \label{fig:roc} 
\end{figure}

In this section, we conduct MeshFace verification experiments on two datasets, i.e., SYN500 and SV1000. Recovered ID photos are used for FVBID according to the aforementioned protocol.
We report ROC curves in Fig.~\ref{fig:roc}. TPR@FPR=1\% (true positive rate when false positive rate is \%1), TPR@FPR=0.1\% and TPR@FPR=0.01\% are reported in Table~\ref{tab:psnr} for closer inspection. We also present the face verification performances with ground-truth clear ID photos and MeshFaces (denoted as \textit{Clear} and \textit{Corrupted}) for fair comparison.

As expected, when using the MeshFaces for verification, the accuracy suffers a severe drop on both datasets due to large detection and alignment errors. After processing the MeshFace with blind face inpainting models, face verification performance with the recovered ID photos has seen a great improvement. Owning to the deeper FCN architecture and expanded receptive fields, all the FCN based models outperform the baseline model MtNet~\cite{7550058} by a large margin on both datasets.

As shown in Fig.~\ref{fig:roc}, feature loss based models (DeMeshNet, FCNF) perform better than the models that only seek a visually pleasing inpainting results (FCNE, FCNW). This suggests that by enforcing a feature level similarity during training, predictions from DeMeshNet lie in a low-dimensional feature space that is closer to the ground-truth clear ID photos than predictions from pixel-level only networks. This is validated in the next section where we calculate the RMSE (rooted mean square error) between the features of ground-truth clear IDs and recovered IDs.


We further investigate the role of the spatial transformer module in our model by comparing DeMeshNet with FCNF which uses the image of original size ($220\times 178$) as input to the feature loss component.
We find that DeMeshNet consistently performs better than FCNF. This is because FCNF takes the whole image, which is different from the aligned face region in both scale and orientation, for feature extraction. Unlike the ImageNet~\cite{russakovsky2015imagenet} models that are trained with multi-scale natural images, the pre-trained face model $\phi$ only takes single-scale, well-aligned face regions for training. The scale and orientation differences have led to the performance degeneration. Therefore, it is significant to take face alignment into account when optimizing the feature level loss as done in DeMeshNet.


It should be noted that for blind inpainting models, there is an upper limit for their verification performance, which is the verification performance with ground-truth clear ID photos. From Table~\ref{tab:psnr}, we can observe that the gap between DeMeshNet and clear ID at TPR@FPR=1\% is very small ($1.4\%$ for both datasets), validating the outstanding performance of DeMeshNet.


\subsection{Evaluation of Inpainting Results} 
\begin{figure*}
  \centering
    \includegraphics[scale=0.63]{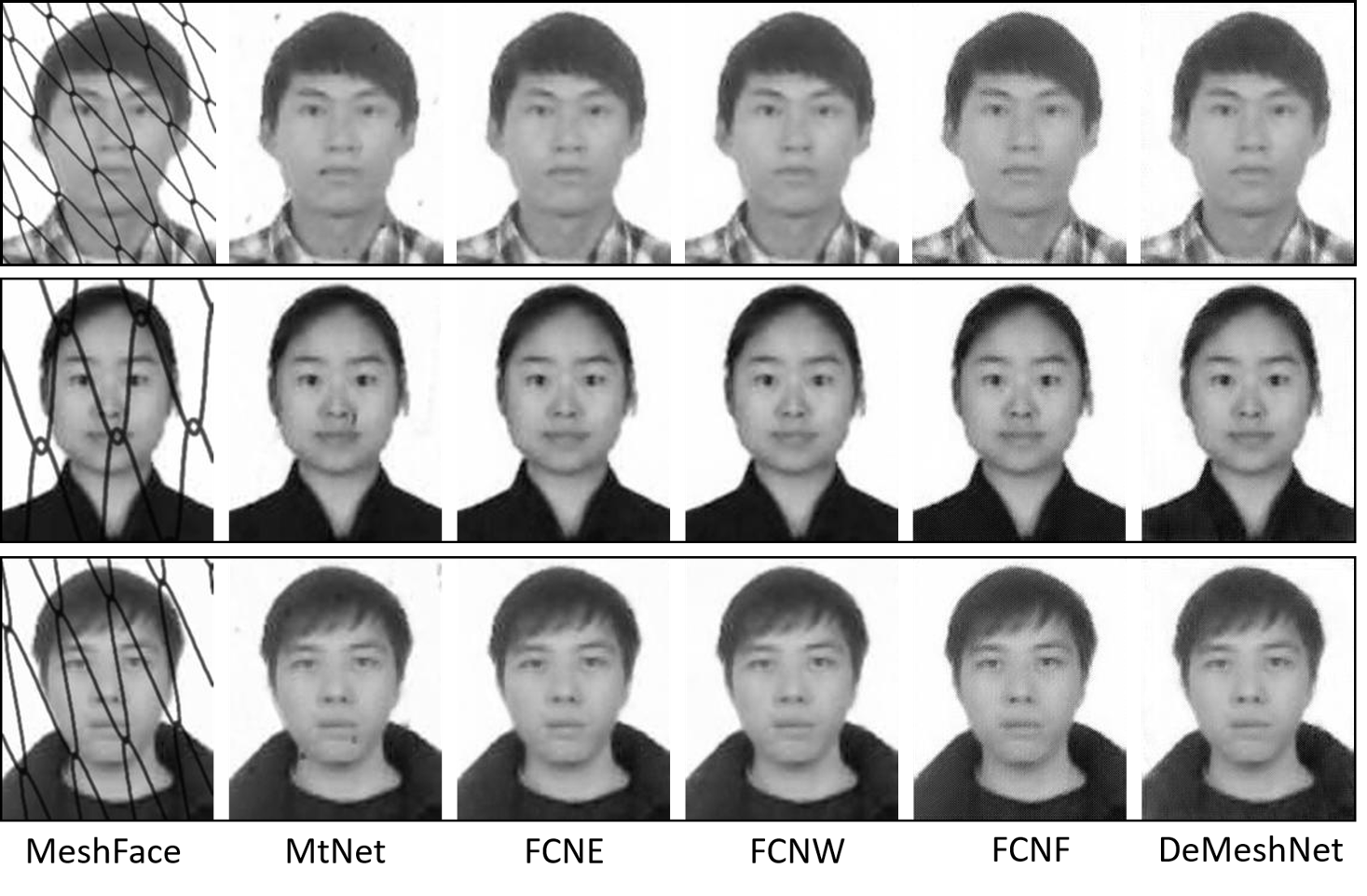}
    \caption{Visual inspection of the inpainting results. Although these inpainting results all look very well and are quite similar, they will lead to entirely different verification rates because of the different RMSE in the feature space.}
    \label{fig:visual} 
\end{figure*}

In this section, we \textit{qualitatively} and \textit{quantitatively} evaluate the inpainting results on the testing set ($200$ individuals, $10000$ photos).
Firstly, we qualitatively evaluate the compared models by visual inspection of the inpainting results in Fig.~\ref{fig:visual}. It is observed that MtNet fails to identify and recover some portions of the corruptions in these cases (cherry-picked to illustrate the point). In contrast, FCN based models can handle all the corruption areas very well because they can enclose more contextual information with expanded receptive fields. This demonstrates the improved capacity of FCN over SRCNN based architectures.

Regarding the details of the recovered ID photo, the models trained with only pixel level loss (FCNE, FCNW, MtNet) can better preserve the consistency of pixels and thus provide a smooth and clear photo which is more similar to the ground-truth. But the images recovered with models trained on feature level loss (FCNF in particular) contain many artifacts, making them visually less appealing. This is because the high-level features are robust to pixel level changes in the texture, shape and even color. However, images recovered from DeMeshNet looks much better than the ones from FCNF due to the introduction of the spatial transformer module within DeMeshNet.



Next, we quantitatively evaluate the models by measuring both pixel level and feature level ($eltwise\_6$) difference. The ground-truth clear ID photo is used as a baseline. Specifically, average PSNR and Euclidean distance between features (or rooted mean square error, RMSE) are shown in Table~\ref{tab:psnr}. Pixel level loss based models yield better PSNR as they explicitly optimize PSNR in their loss function. We also observe that FCNW performs slightly better than FCNF. This implies that exploiting extra supervision from the corruption position in the training phase is beneficial for acquiring visually pleasing inpainting results.

To validate the choice of the reverse Huber loss over the Euclidean loss on the feature level difference, we also implement a DeMeshNet\_E that use Euclidean loss for the feature level loss. From Table~\ref{tab:psnr}, we can see that RMSE for DeMeshNet\_E is slightly larger than DeMeshNet. This indicates that the reverse Huber loss is more effective at minimizing the feature level distance thanks to the imposed L1 norm when residuals are small.

Furthermore, from Table~\ref{tab:psnr} we observe that smaller RMSE often means better verification performance, but higher PSNR doesn't guarantee smaller RMSE. This reveals that the models trained with only pixel level loss does suffer from the influence of the easily fooled nature of CNN~\cite{goodfellow2014explaining,nguyen2015deep,szegedy2013intriguing}. Moreover, visually appealing inpainting results does not necessarily produce better verification results. By exploring supervision from the deep feature space, DeMeshNet can capture a distribution that is more robust to transformation in $\phi$ and thus provide a stable high-level representation in the compact deep feature space.

\section{Conclusions}

This paper addresses the MeshFace verification problem that verifies corrupted ID photos against clear daily photos. Specifically, we have proposed DeMeshNet that consists of three parts to blindly inpaint the MeshFace before conducting verification.The proposed DeMeshNet distinguishes itself from previous works by explicitly taking verification performance into consideration while recovering a clear ID photo. The training objective of DeMeshNet is motivated by the fact that minimizing pixel level differences alone cannot guarantee a small intra-class feature distance in the compact deep feature space, which is crucial for accurate face verification. By further incorporating a spatial transformer module, DeMeshNet can implement face alignment within the network, resulting in an end-to-end network. For optimizing DeMeshNet, a very well-performed facial feature extraction network has been trained in advance. Experimental results on two MeshFace datasets demonstrate that the proposed DeMeshNet outperforms previous work on verification performance.

{\small
\bibliographystyle{ieee}
\bibliography{egbib}
}

\end{document}